\newcommand{\Eq}[1]{\begin{equation}#1\end{equation}}
\newcommand{\Vc}[1]{\mbox{\boldmath$#1$}}
\newcommand{\ds}[1]{\mathrm{d}{#1}}
\newcommand{\dv}[1]{\mathrm{d}\Vc{#1}}

\newcommand{\mR}{\mathbb{R}}
\def\Gg{G}
\newcommand{\mM}{\mathcal{M}}

\DeclareMathAlphabet{\mathsfsl}{OT1}{cmss}{m}{sl}
\newcommand{\tensor}[1]{\mathsfsl{#1}}

\documentclass[twocolumn,10pt]{article}

\usepackage{pifont,calc,amsfonts,amsbsy,amssymb,tabularx,float,times,epsf}
\usepackage{hero}

\lineskip
2pt \normallineskip     1pt

\begin{document}

\title{Manifold Learning with Geodesic Minimal Spanning Trees}
    \author{Jose Costa and Alfred Hero \\
University of Michigan, Ann Arbor, MI 48109-2122}
\maketitle

\bc
\bf Abstract
\ec

{\it In the manifold learning problem one seeks to discover a smooth
low dimensional surface, i.e., a manifold embedded in a higher
dimensional linear vector space, based on a set of measured sample
points on the surface. In this paper we consider the closely
related problem of estimating the manifold's intrinsic dimension
and the intrinsic entropy of the sample points. Specifically, we
view the sample points as realizations of an unknown multivariate
density supported on an unknown smooth manifold. We present a
novel geometrical probability approach, called the
geodesic-minimal-spanning-tree (GMST), to obtaining asymptotically
consistent estimates of the manifold dimension and the R\'{e}nyi
$\alpha$-entropy of the sample density on the manifold. The GMST
approach is striking in its simplicity and does not require
reconstructing the manifold or estimating the multivariate density
of the samples. The GMST method simply constructs a minimal
spanning tree (MST) sequence using a geodesic edge matrix and uses
the overall lengths of the MSTs to simultaneously estimate
manifold dimension and entropy.
We illustrate the GMST approach for dimension and
entropy estimation of a human face dataset.}


\section{Introduction}

Consider a class of natural occurring signals, e.g., recorded
speech, audio, images, or videos.  Such signals typically have
high extrinsic dimension, e.g., as characterized by the number of
pixels in an image or the number of time samples in an audio
waveform.  However, most natural signals have smooth and regular
structure, e.g. piecewise smoothness, that permits substantial
dimension reduction with little or no loss of content information.
For support of this fact one needs only consider the success of
image, video and audio compression algorithms, e.g. MP3, JPEG and
MPEG, or the widespread use of efficient computational geometry
methods for rendering smooth three dimensional shapes.

A useful representation of a regular signal class is to model it
as a set of vectors which are constrained to a smooth low
dimensional manifold embedded in a high  dimensional vector space.
This manifold may in some cases  be a linear, i.e., Euclidean,
subspace but in general it is a non-linear curved surface. A
problem of substantial recent interest in machine learning,
computer vision, signal processing and statistics
\cite{Tenenbaum&etal:Science00,Silva&Tenenbaum:02,Kirby:01,Donoho&Grimes:TR03,Kegl:NIPS02,Verveer&Duin:PAMI95}
is the determination of the so-called intrinsic dimension of the
manifold and the reconstruction of the manifold from a set of
samples from the signal class. This problem falls in the area of
manifold learning which is concerned with discovering low
dimensional structure in high dimensional data.

When the samples are drawn from a large population of signals one
can interpret them as realizations from a  multivariate
distribution supported on the manifold. As this distribution is
singular in the higher dimensional embedding space it has zero
entropy as defined by the standard Lebesgue integral over the
embedding space. However, when defined as a Lebesgue integral
restricted to the lower dimensional manifold the entropy can be
finite. This finite ``intrinsic" entropy can be useful for for
exploring data compression over the manifold or, as suggested in
\cite{Hero&etal:SpMag02}, clustering of multiple sub-populations
on the manifold. The question that we address in this paper is:
how to simultaneously estimate the intrinsic dimension and
intrinsic entropy on the manifold given a set of random sample
points? We present a novel geometrical probability  approach to
this question which is based on entropic graph methods developed
by us and reported in publications
\cite{Hero&Michel:IT99,Hero&etal:SpMag02,Hero&etal:IT02}.

Techniques for manifold learning can be classified into three
categories: linear methods, local methods, and global methods.
Linear methods include principal components analysis (PCA)
\cite{Jain&Dubes:88} and multidimensional scaling (MDS)
\cite{Cox&Cox:94}. They are based on analyzing eigenstructure of
empirical  covariance matrices, and can be reliably applied only
when the manifold is a linear subspace. Local methods include
linear local imbedding (LLE) \cite{Roweis&Saul:Science00}, locally
linear projections (LLP) \cite{Huo&Chen:GENSIPS02}, Laplacian
eigenmaps \cite{Belkin&Niyogi:NIPS02}, and Hessian eigenmaps
\cite{Donoho&Grimes:TR03}. They are based on local approximation
of the geometry of the manifold, and are computationally simple to
implement.  Global approaches include ISOMAP
\cite{Tenenbaum&etal:Science00} and C-ISOMAP
\cite{Silva&Tenenbaum:03}. They preserve the manifold geometry at
all scales, and have better stability than local methods.

We propose a geodesic-minimal-spanning-tree (GMST) method for
manifold learning that is implemented as follows. First a complete
geodesic graph between all pairs of data samples is constructed,
e.g. using ISOMAP or C-ISOMAP. Then a minimal spanning graph, the
GMST, is obtained by pruning the complete geodesic graph down to a
subgraph that still connects all points but has minimum total
geodesic length. The intrinsic dimension and intrinsic
$\alpha$-entropy is then estimated from the GMST length functional
using a simple linear least squares (LLS) and method of moments
(MOM) procedure.

The GMST method falls in the category of global approaches to
manifold learning but it differs significantly from the
aforementioned methods. First, it has a different scope. Indeed,
unlike ISOMAP and C-ISOMAP, the GMST method provides a
statistically consistent estimate of the intrinsic entropy in
addition to the intrinsic dimension of the manifold. To the best
of our knowledge no other such technique has been proposed for
learning manifold dimension. Second, unlike local methods that
work on chunks of data in local neighborhoods, GMST works on
chunks of resampled data over the global data set. Third, for $N$
samples the GMST method has $O(N \log N)$ computational complexity
as compared with the $O(N^3)$ complexity of an MDS ISOMAP
reconstruction. Fourth, the GMST method is simple and elegant: it
estimates intrinsic entropy and dimension by detecting the rate of
increase of a GMST as a function of the number of its resampled
vertices.

The aims of this paper are limited to introducing GMST as a novel
method for estimating manifold dimension and entropy of the
samples. As in work of others on dimension estimation
\cite{Kegl:NIPS02,Camastra&Vinciarelli:PAMI02} we do not here
consider the issue of reconstruction of the complete manifold.
Similarly to these authors, we believe that dimension estimation
and entropy estimation for non-linear data are of interest in
their own right. We also do not consider the effect of additive
noise or outliers on the performance of GMST. Finally, the
consistency results of GMST reported here are limited to domain
manifolds defined by some smooth unknown mapping. The extension of
GMST methodology to general target manifolds, e.g. those defined
by implicit level set embeddings
\cite{Memoli&etal:TR02,Memoli&Sapiro:JCP01}, is a worthwhile topic
for future investigation.

What follows is a brief outline of the paper. We review some
necessary background on the mathematics of domain manifolds in
Sec. \ref{S:Background}. In Sec. \ref{S:entropic} we review the
asymptotic theory of entropic graphs and obtain several new
results required for their extension to embedded manifolds.  In
Sec. \ref{S:GMST} we define the general GMST algorithm. Finally in
Sec. \ref{S:Application} we illustrate the GMST approach to
estimating intrinsic dimension and entropy of a human face
dataset.

\section{Background}
\label{S:Background}

\subsection{A 3D Example}

To illustrate ideas consider a 2D surface embedded in 3D Euclidean
space, called the embedding space. %
Let $\{\Vc{x}_1, \Vc{x}_2, \ldots\} \subset U \subseteq \mR^2$ be
a set of points (samples) in a subset $U$ of the plane. Naturally,
the shortest path between any pair $(\Vc{x}_i, \Vc{x}_j)$ of these
points is given by the straight line in $\mR^2$ connecting them,
with corresponding distance given by its Euclidean  ($L_2$)
length, $|\Vc{x}_i- \Vc{x}_j|_2$. Now let $U$ be used as a
parameterization space to describe a curved surface in $\mR^3$ via
a mapping $\varphi:U \rightarrow \mR^3$. Surfaces $\mM=\varphi(U)$
defined in this explicit manner are called domain manifolds and
they inherit the topological dimension, equal to 2 in this case,
of the parameterization space. When $\varphi$ is non-linear the
shortest path on $\mM$ between points $\Vc{y}_i=\varphi(\Vc{x}_i)$
and $\Vc{y}_j=\varphi(\Vc{x}_j)$ is a curve on the surface called
the geodesic curve. In this paper we will primarily consider
domain manifolds defined by conformal mappings $\varphi$. Such
conformal embeddings have the property that the length of paths on
the surface are identical to lengths of paths in the
parameterization space, possibly up to a smoothly varying local
scale factor. This property guarantees that, regardless of how the
mapping $\varphi$ ``deforms" $U$ onto $\mM$, the geodesic
distances in $\mM$ are closely related to the Euclidean distances
in $U$.  When this smooth surface representation holds there exist
algorithms, e.g. ISOMAP and C-ISOMAP
\cite{Tenenbaum&etal:Science00,Silva&Tenenbaum:03}, which can be
used to estimate the Euclidean distances between points in $U$
from estimates of the geodesic distances between points in $\mM$.
If a certain type of minimal spanning graph is constructed using
these estimates well established results in geometrical
probability \cite{Yukich:98,Hero&etal:SpMag02} allow us to develop
simple estimates of both entropy and dimension of the points
distributed on the surface.

\subsection{Differential Geometry Setting}

In the following, we recall some facts from differential geometry
needed to formalize and generalize the ideas just described. We
will consider smooth manifolds embedded in $\mR^d$. For the
general theory we refer the reader to any standard book in
differential geometry (for example, \cite{Carmo:76},
\cite{Carmo:92}, \cite{Boothby:03}). An $m$-dimensional
\emph{smooth manifold} $\mM \subseteq \mathbb{R}^d$ is a set such
that each of its points has a neighborhood that can be
parameterized by an open set of $\mR^m$ through a local change of
coordinates. Intuitively, this means that although $\mM$ is a
(hyper) surface in $\mathbb{R}^d$, it can be locally identified
with $\mR^m$.

Let $\varphi:\Omega \mapsto \mM$ be a mapping between two
manifolds, $\Omega, \mM$. Let $\gamma$ be a curve in $\Omega$. The
\emph{tangent map} $\ds{\varphi}_{\Vc{x}}$ assigns each tangent
vector $\Vc{v}$ to $\Omega$ at point $\Vc{x}$ the tangent vector
$\ds{\varphi}_{\Vc{x}}\Vc{v}$ to $\mM$ at point $\varphi(\Vc{x})$,
such that, if $\Vc{v}$ is the initial velocity of $\gamma$ in
$\Omega$, then $\ds{\varphi}_{\Vc{x}} \Vc{v}$ is the initial
velocity of the curve $\varphi(\gamma)$ in $\mM$. For example, if
$\Vc{x} \in U \subseteq \Omega \subseteq \mR^m$, with $U$ an open
set of $\mR^m$, then $\ds{\varphi}_{\Vc{x}}\Vc{v} =
\tensor{J}_\varphi(\Vc{x}) \, \Vc{v}$, where $\tensor{J}_\varphi
=[\partial \varphi_i / \partial x_j]$, $i=1,\ldots,d$,
$j=1,\ldots,m$, is the Jacobian matrix associated with $\varphi$
at point $\Vc{x} \in \Omega$.

The \emph{length} of a smooth curve $\Gamma:[0,1] \mapsto \mM$ is
defined as $\ell(\Gamma)=\int_0^1 |\dot{\Gamma}(t)| \ds{t}$. The
\emph{geodesic distance} between points $\Vc{y}_0,\Vc{y}_1 \in
\mM$ is the length of the shortest (piecewise) smooth curve
between the two points:
    \[
        d_{\mM}(\Vc{y}_0,\Vc{y}_1) = \inf_\Gamma\{\ell(\Gamma): \Gamma(0)=\Vc{y}_0, \Gamma(1)=\Vc{y}_1\} \ .
    \]

We can now define the following types of embeddings.

\begin{definitions}  \label{D:ConformalMapping}
$\varphi:\Omega \mapsto \mM$ is called a conformal mapping if
$\varphi$ is a diffeomorphism (i.e., $\varphi$ is differentiable,
bijective with differentiable inverse $\varphi^{-1}$) and, at each
point $\Vc{x} \in \Omega$, $\varphi$ preserves the angles between
tangent vectors, i.e.,
    \Eq{
        \left( \ds{\varphi}_{\Vc{x}} \Vc{v} \right)^T \left( \ds{\varphi}_{\Vc{x}} \Vc{w} \right) = c(\Vc{x}) \, \Vc{v}^T \Vc{w} \ ,
    }
for all vectors $\Vc{v}$ and $\Vc{w}$ that are tangent to $\Omega$
at $\Vc{x}$, and $c(\Vc{x})>0$ is a scaling factor that varies
smoothly with $\Vc{x}$. If for all $\Vc{x} \in \Omega$,
$c(\Vc{x})=1$, then $\varphi$ is said to be a (global) isometry.
In this case the length of tangent vectors is also preserved in
addition to the angles between them.
\end{definitions}

It is easy to check that if there is an open set $U \subseteq
\Omega \subseteq \mR^m$, then the diffeomorphism $\varphi$ is a
conformal mapping iff $\tensor{J}_\varphi(\Vc{x})^T
\tensor{J}_\varphi(\Vc{x}) = c(\Vc{x}) \, \tensor{I}_{m}$, where
$\tensor{I}_{m}$ is the $m \times m$ identity matrix. In this
case, the geodesic distance in $\mM$ can be computed as follows.
Any smooth curve $\Gamma:[0,1] \mapsto \mM$ can be represented as
$\Gamma(t) = \varphi(\gamma(t))$, where $\gamma:[0,1] \mapsto
\Omega$ is a smooth curve in $\mR^m$. Then, the length
$\ell(\Gamma)$ of the curve $\Gamma$ is given by
\ben
        \ell(\Gamma) &=& \int_0^1 \left| \frac{\ds{}}{\ds{t}}
        \varphi(\gamma(t)) \right| \ds{t}
\\
        &=& \int_0^1 |\tensor{J}_\varphi(\gamma(t)) \, \dot{\gamma}(t)| \, \ds{t} = \int_0^1 \sqrt{c(\gamma(t))} \, |\dot{\gamma}(t)| \, \ds{t} \ .
\een
As in $\mR^m$ the shortest path between any two points is given by
the straight line that connects them, $\gamma(t)= \Vc{x}_0 + t
(\Vc{x}_1-\Vc{x}_0)$ minimizes $\int_0^1  |\dot{\gamma}(t)| \,
\ds{t}$, over all smooth curves with start and end points at
$\Vc{x}_0$ and $\Vc{x}_1$, respectively. So, if $c(\Vc{x})= c$,
for all $x \in \Omega$, the geodesic distance between
$\Vc{y}_0=\varphi(\Vc{x}_0)$ and $\Vc{y}_1=\varphi(\Vc{x}_1)$ is
\Eq{ \label{E:GeodesicDistance}
d_{\mM}(\varphi(\Vc{x}_0),\varphi(\Vc{x}_1))= c
|\Vc{x}_0-\Vc{x}_1|_2 \ .}
When $c=1$, i.e., $\varphi$ is an isometry, the geodesic distance
in $\mM$ and the Euclidean distance in the parameterization space
$\mR^m$ are the same. If $c>1$ ($c<1$) there is a global expansion
(contraction) in the distances between points.

It is evident from the above discussion that geodesic distances
carry strong information about a non-linear domain manifold such
as $\mM$. However, their computation requires the knowledge of the
analytical form of $\mM$ via $\varphi$ and its Jacobian. In the
manifold learning scenario considered in this paper this
analytical form is assumed unknown and, instead, we are given a
finite set of data points lying on the smooth $m$-dimensional
manifold $\mM$, with $m$ also considered unknown. In order to
reconstruct a domain manifold along with its parameterization we
need to estimate the geodesic distances between pairs of data
points in $\mM$ and the respective Euclidean distances betweem
pre-images of these data points in the parameterization space $U$.

When $\mM$ is an isometric embedding the ISOMAP algorithm
\cite{Tenenbaum&etal:Science00} obtains such a reconstruction from
a finite set of samples through estimation of the pairwise
geodesic distances. This estimate is computed from a Euclidean
graph $\Gg$ connecting all local neighborhoods of data points in
$\mM$. Specifically, ISOMAP proceeds as follows. Two methods,
called the $\epsilon$-rule and the $k$-rule
\cite{Tenenbaum&etal:Science00}, have been proposed for
contructing $\Gg$. The first method connects each point to all
points within some fixed radius $\epsilon$ and the other connects
each point to all its $k$-nearest neighbors. The graph $\Gg$
defining the connectivity of these local neighborhoods is then
used to   approximate the geodesic distance between any pair of
points as the shortest path through $\Gg$ that connects them. This
results in an edge matrix whose $(i,j)$ entry is the geodesic
distance estimate for the $(i,j)$-th pair of points. Finally,
ISOMAP obtains a smooth reconstruction of the manifold by applying
the classical Multidimensional Scaling (MDS) method
\cite{Cox&Cox:94} to the edge matrix.

\begin{table}
\begin{center}
\begin{tabular}{|lp{6cm}|}
\hline Step 1. & Determine a Euclidean neighborhood graph $\Gg$ of
the observed data $\mathcal{Y}_n$ according to the $\epsilon$-rule
or
the $k$-rule as defined in ISOMAP \cite{Bernstein&etal:00}. \\
[1ex] Step 2. & For isometric embeddings compute the edge matrix
$\calE$ of the ISOMAP graph \cite{Tenenbaum&etal:Science00}  and
for conformal imbeddings compute the edge matrix $\calE$ of the
C-ISOMAP graph \cite{Silva&Tenenbaum:03}. The $(i,j)$ entry of
this symmetric matrix is the sum of the lengths of the edges in
$\Gg$ along the shortest path between the pair of vertices
$(\Vc{Y}_i,\Vc{Y}_j)$ where the edge lengths between neighboring
points $\Vc{Y}_1$, $\Vc{Y}_2$ in $\Gg$ are defined as Euclidean
distance $|\Vc{Y}_1$-$\Vc{Y}_2|$ in the case of ISOMAP or
$|\Vc{Y}_1$-$\Vc{Y}_2|/\sqrt{M(1)M(2)}$ in the case of C-ISOMAP
where $M(i)$ is the mean distance of $\Vc{Y}_i$ to its immediate
nearest neighbors.
\\
\hline
\end{tabular}
\caption{First two steps of the ISOMAP/C-ISOMAP algorithms to
reconstruct Euclidean distances between $\calX_n$ on the embedding
parameterization space from points $\calY_n$ over the embedded
manifold} \label{Tab:DistancesEstimate}
\end{center}
\end{table}

Steps one and two of ISOMAP are motivated by the fact that locally
any smooth manifold is approximately ``flat" and, so, the
distances between neighboring points are well approximated by
their Euclidean distances. For faraway points, the geodesic
distance is estimated by summing the sequence of such local
approximations over the shortest path through the graph $\Gg$. In
\cite{Bernstein&etal:00} it was proved that, when the data are
random samples from a continuous distribution on the manifold
$\mM$, the first two steps of ISOMAP recover the true geodesic
distances with high probability if the data points form a
sufficiently ``dense" sampling of $\mM$ and if $\mM$ is free of
``holes." When $\mM$ is a global isometric embedding in $\mR^d$,
the estimated geodesic distances are also an estimate of distances
in $\mR^m$ and the ISOMAP succeeds in its task of manifold
reconstruction. For other types of embeddings, there is no
guarantee that the ISOMAP will recover the correct
parameterization. In \cite{Silva&Tenenbaum:02}, a variant of this
algorithm, called C-ISOMAP, was proposed to deal with the more
general class of conformal embeddings.

With regards to estimation of the intrinsic dimension $m$ several
methods have been proposed \cite{Jain&Dubes:88}. Most of these
methods are based on linear projection techniques: a linear map is
explicitly constructed and dimension is estimated by applying
Principal Component Analysis (PCA), factor analysis, or MDS to
analyze the eigenstructure of the data. These methods rely on the
assumption that only a small number of the eigenvalues of the
(processed) data covariance will be significant.  Linear methods
tend to overestimate $m$ as they don't account for non-linearities
in the data. Both nonlinear PCA \cite{Kirby:01} methods and the
ISOMAP circumvent this problem but they still rely on unreliable
and costly  eigenstructure estimates. Other methods have been
proposed based on local geometric techniques, e.g., estimation of
local neighborhoods \cite{Verveer&Duin:PAMI95} or fractal
dimension \cite{Camastra&Vinciarelli:PAMI02}, and estimating
packing numbers  \cite{Kegl:NIPS02} of the manifold.

\section{Entropic Graph Estimators on Embedded Manifolds}
\label{S:entropic}

Let $\calY_n=\Vc{Y}_1,\ldots, \Vc{Y}_n$ be $n$ independent
identically distributed (i.i.d.) random vectors in $[0,1]^d$, with
multivariate Lebesgue density $f$, which we will also call random
vertices. Define the edge matrix $\calE$ as the $n \times n$
matrix of edge lengths (w.r.t. a specified metric) between pairs
of vertices. A spanning graph $T$ over $\calY_n$ is defined as the
pair $\{V,E\}$ where $V = \calY_n$ and $E$ is a subset of edges
from $\calE$ which connect the vertices $V$. When $\calE$ is
computed from pairwise Euclidean distances $T$ is called a
Euclidean spanning graph.

It has long been known \cite{Beardwood&etal:CambPhil59} that, when
suitably normalized, the sum of the edge weights of certain
minimal Euclidean spanning graphs $T$ over $\calY_n$ converges
almost surely (a.s.) to the limit $\beta_{d}\int_{\mR^d}
f^{\alpha}(\Vc{y}) d\Vc{y}$ where  where the integral is
interpreted in the sense of Lebesgue, $\alpha \in (0,1)$ and
$\beta_{d}>0$.
This a.s. limit is the integral factor $\int f^\alpha$ in what we
will call the {\it extrinsic} R\'{e}nyi $\alpha$-entropy of the
multivariate Lebesgue density $f$:
\be H_{\alpha}^{\mR^d}(f)=\frac{1}{1-\alpha} \log \int_{\mR^d}
f^{\alpha}(\Vc{y}) d\Vc{y} \ . \label{eq:extent} \ee
  In the limit, when $\alpha \to 1$ we obtain the usual Shannon
entropy, $-\int_{\mR^d} f(\Vc{y}) \log f(\Vc{y}) \dv{y}$. Graph
constructions that converge to the integral in the limit
(\ref{eq:extent}) were called continuous quasi-additive
(Euclidean) graphs in \cite{Yukich:98} and entropic (Euclidean)
graphs in \cite{Hero&etal:SpMag02}.
See the monographs by Steele \cite{Steele:97} and Yukich
\cite{Yukich:98} for an excellent introduction to the theory of
such random Euclidean graphs.

The $\alpha$-entropy has proved to be an important quantity in
signal processing, where its applications range from vector
quantization \cite{Gersho:IT79,Neuhoff:IT96} to pattern matching
\cite{Hero&Michel:HOS99} and image registration
\cite{Hero&etal:SpMag02,Heemuchwala&etal:EURASIP02}. The
$\alpha$-entropy parameterizes the Chernoff exponent governing the
minimum probability of error  \cite{Cover&Thomas:91} making it an
important quantity in detection and classification problems. Like
the Shannon entropy, the $\alpha$-entropy also has an operational
characterization in terms of source coding rates. In
\cite{Csiszar:IT95} it was shown that the $\alpha$-entropy of a
source determines the achievable block-code rates in the sense
that the probability of block decoding error converges to zero at
an exponential rate with rate constant $H_\alpha^{\mR^d}(f)$.

\subsection{Beardwood-Halton-Hammersley Theorem in $\mR^d$}

A remarkable result in geometrical probability was  established by
Beardwood, Halton and Hammersley almost half a century ago
\cite{Beardwood&etal:CambPhil59}. Let $\mathcal{Y}_n = \{
\Vc{Y}_1, \ldots, \Vc{Y}_n \}$ be a set of points in $\mR^d$.
A minimal Euclidean graph spanning $\calY_n$ is defined as the
graph spanning $\calY_n$ having minimal overall length
    \Eq{ \label{E:OverallLenght1}
    L^{\mR^d}_{\gamma}(\calY_n) = \min_{T \in \mathcal{T}} \sum_{e \in T} |e|^\gamma  \ . }
Here the sum is over all edges $e$ in the graph $T$, $|e|$ is the
Euclidean length of $e$, and $\gamma \in (0,d)$ is called the
\emph{edge exponent} or \emph{power-weighting constant}. For
example when $\mathcal{T}$ is the set of spanning trees over
$\mathcal{Y}_n$ one obtains the MST. A minimal Euclidean graph is
continuous quasi-additive when it satisfies several technical
conditions specified in \cite{Yukich:98} (also see
\cite{Hero&Michel:IT99}). Continuous quasi-additive Euclidean
graphs include: the minimal spanning tree (MST), the $k$-nearest
neighbors graph ($k$-NNG), the minimal matching graph (MMG), the
traveling salesman problem (TSP), and their power-weighted
variants.  While all of the results in this paper apply to this
larger class of minimal graphs we specialize to the MST for
concreteness.

{\bf Beardwood-Halton-Hammersley (BHH) Theorem
\cite{Steele:97,Yukich:98}}: {\it Let $\calY_n$ be an i.i.d. set
of random variables taking values in $\mR^d$ having common
probability distribution $P$. Let this distribution have the
decomposition $P=F+Q$ where $F$ is the Lebesgue continuous
component and $Q $ is the singular component. The Lebesgue
continuous component has a Lebesgue density (no delta functions)
which is denoted $f(x)$, $x\in \mR^d$. Let
$L^{\mR^d}_\gamma(\calY_n)$ be the length of the MST spanning
$\calY_n$ and assume that $d \geq 2$ and $0< \gamma < d$. Then
\be L^{\mR^d}_{\gamma}(\mathcal{Y}_n)/n^{\alpha} \rightarrow
\beta_{d}\int_{\mR^d} f^{\alpha}(\Vc{y}) d\Vc{y} \label{eq:Linf}
\;\;\;\ (a.s.), \ee
where $\alpha=(d-\gamma)/d$ and $\beta_{d}$ is a constant not
depending on the distribution $P$.
Furthermore, the mean length
$E[L^{\mR^d}_\gamma(\calY_n)]/n^{\alpha}$ converges to the same
limit.}

The limit on the right side of (\ref{eq:Linf}) in the BHH theorem
is zero when the distribution $P$ has no Lebesgue continuous
component, i.e., when $F\equiv 0$. On the other hand, when $P$ has
no singular component, i.e., $Q\equiv 0$, a consequence of the BHH
Theorem is that
 \Eq{ \label{E:EuclideanConvergence}
\hat{H}_\alpha^{\mR^d}(\calY_n) \defined \frac{d}{\gamma} \left[
\log \frac{L^{\mR^d}_{\gamma}(\mathcal{Y}_n)}{n^{(d-\gamma)/d}}
-\log \beta_d\right]
    }
is an asymptotically unbiased and strongly consistent estimator of
the extrinsic $\alpha$-entropy $H_\alpha^{\mR^d}(f)$ defined in
(\ref{eq:extent}).

\subsection{Generalization of BHH Thm. to Embedded Manifolds}

If the vertices $\calY_n=\{\Vc{Y}_1,\ldots, \Vc{Y}_n\}$  are
constrained to lie on a smooth $m$-dimensional manifold $\mM
\subset [0,1]^d$, the distribution of $\Vc{Y}_i$ is singular with
respect to Lebesgue measure, $F\equiv 0$, and, as previously
mentioned, the limit (\ref{eq:Linf}) in the BHH Theorem is zero.
However, as shown below, if $\mM$ is defined by an isometric
embedding from the parameterization space $\mR^m$, if $\Vc{Y}_i$
has a continuous density $f$ on $\mM$, and if ISOMAP is used to
approximate the geodesic edge
matrix, %
then the length of an MST  constructed from the geodesic edge
matrix can be made to converge, after suitable normalization and
transformation, to the {\it intrinsic} $\alpha$-entropy
$H_\alpha^{\mM}(f)$ on $\mM$ defined by
\be \label{eq:intent} H_\alpha^{\mM}(f)=\frac{m}{\gamma}\log
\int_{\mM} f^{\alpha}(\Vc{y})  \mu_{\mM}(d\Vc{y}), \ee
where $\mu_{\mM}(d\Vc{y})$ denotes the differential volume element
over $\mM$.

More generally, assume that $\mM$ is embedded in $[0,1]^d$ through
the diffeomorphism $\varphi$. %
As $\Vc{X}_i= \varphi^{-1}(\Vc{Y}_i)$ lives in $\mR^m$, let $T$ be
the Euclidean minimal graph spanning $\calX_n$ and having length
function $ L^{\mR^m}_{\gamma} \left( \mathcal{X}_n
\right)=L^{\mR^m}_{\gamma} \left( \varphi^{-1}(\mathcal{Y}_n)
\right)$ according to definition (\ref{E:OverallLenght1}).
We have the following extension of the BHH Theorem.

\begin{theorems} \label{T:ManifoldGraph}
    Let $\mM$ be a smooth compact $m$-dimensional manifold embedded in
    $[0,1]^d$  through the diffeomorphism $\varphi : \mR^m \mapsto \mM$.
    Assume $2 \leq m \leq
    d$ and $0 < \gamma < m$.
Suppose that $\Vc{Y}_1, \Vc{Y}_2, \ldots$ are i.i.d.\ random
vectors on $\mM$ having common density $f$ with respect to
Lebesgue measure $\mu_{\mM}$ on $\mM$. Then, the length functional
$L^{\mR^m}_\gamma(\varphi^{-1}(\calY_n))$ of the MST  spanning
$\varphi^{-1}(\calY_n)$ satisfies
\be
&&\lim_{n \to \infty}
L^{\mR^m}_\gamma(\varphi^{-1}(\calY_n))/n^{(d^{'}-\gamma)/d^{'}} 
        \to \label{eq:Linf2} \\ 
&&\left\{ \begin{array}{cc} \infty , & d^{'} < m \\ \\
        \beta_{m} \int_{\mM}
        \left[ \det \left( \tensor{J}_\varphi^T \tensor{J}_\varphi \right) \right]^{\frac{\alpha-1}{2}}
        f^\alpha(\Vc{y}) \, \mu_{\mM}(d\Vc{y}) , & d^{'} = m \\ \\
        0, & d^{'} > m \end{array} \right. \nonumber
\ee
(a.s.) where $\alpha=(m-\gamma)/m$.  Furthermore, the mean
$E[L^{\mR^m}_\gamma(\varphi^{-1}(\calY_n))]/n^{(d^{'}-\gamma)/d^{'}}$
converges to the same limit.
\end{theorems}

This theorem is a simple consequence of the relation
(\ref{eq:Linf}) in the BHH Theorem and properties of integrals
over manifolds.

\noindent{\it Proof of Thm. \ref{T:ManifoldGraph}:}
By the BHH Theorem, with probability one
\be
L^{\mR^m}_\gamma(\calX_n) = n^{(m-\gamma)/m} \beta_{m} \;
\int_{\mR^m}f^\alpha_X(\Vc{x}) \, d\Vc{x}
        +o(n^{(m-\gamma)/m}),\label{E:BHHp1} \ee
where $f_X$ is the density of $\Vc{X}_i=\varphi^{-1}(\Vc{Y}_i)$.
Therefore the limits claimed in (\ref{eq:Linf2}) for $d^{'}<m$ and
$d^{'}>m$ are obvious. For $d^{'}=m$ the relation (\ref{E:BHHp1})
implies
\be \lim_{n\rightarrow\infty}
L^{\mR^m}_\gamma(\calX_n)/n^{(m-\gamma)/m} = \beta_{m} \;
\int_{\mR^m}f^\alpha_X(\Vc{x}) \, d\Vc{x} , \label{E:BHHp2} \ee
and it remains to show that this limit is identical to the limit
asserted in (\ref{eq:Linf2}).

For an integrable function $F$ defined on a domain manifold $\mM$
defined by the diffeomorphism $\varphi:\mR^m \mapsto \mM$, the
integral of $F$ over $\mM$ satisfies the relation
\cite{Carmo:92}:
    \Eq{ \label{E:ChangeVariable} \int_{\mM} F(\Vc{y}) \, \mu_{\mM}(d\Vc{y}) =
    \int_{\mR^m} F(\varphi(\Vc{x})) \, g(\Vc{x}) \, d\Vc{x} \ , }
where $g(\Vc{x})= \sqrt{\det \left(  \tensor{J}_\varphi^T
\tensor{J}_\varphi \right)}$.  Specializing $F$ to the indicator
function of a small volume centered at a point $\Vc{y}$
(\ref{E:ChangeVariable}) implies the following relation between
volume elements in $\mM$ and $\mR^m$: $
\mu_{\mM}(d\Vc{y})=g(\Vc{x}) \, d\Vc{x} .$  Furthermore,
specializing to $F(\Vc{y})=f(\Vc{y})$ it is clear from
(\ref{E:ChangeVariable}) that $f_X(\Vc{x})=f(\varphi(\Vc{x}))
g(\Vc{x})$.
Therefore
\ben \int_{\mR^m} f_X^\alpha(\Vc{x}) d\Vc{x} &=& \int_{\mR^m}
(f(\varphi(\Vc{x})) g(\Vc{x}))^\alpha d\Vc{x} \\ &=& \int_{\mR^m}
f^\alpha(\varphi(\Vc{x})) g^{\alpha-1}(\Vc{x})) \; g(\Vc{x})
d\Vc{x} , \een
which, after the change of variable $\Vc{x} \mapsto
\varphi(\Vc{x})$, is equivalent to the integral in the limit
(\ref{eq:Linf2}). \qed

Our goal is to learn the entropy of non-linear data on a domain
manifold together with its intrinsic dimension, given only the
data set $\mathcal{Y}_n$ of $n$ samples in the embedding space
$\mR^d$, and without knowledge of its embedding function
$\varphi$. If $\varphi$ is an isometric or conformal embedding
then it has been shown that for sufficiently dense sampling over
$\mM$, i.e., for large $n$, the ISOMAP or the C-ISOMAP algorithm
summarized in Table \ref{Tab:DistancesEstimate} will approximate
the matrix  of pairwise Euclidean distances between the points
$\calX_n=\varphi^{-1}(\calY_n)$ in the domain space $\mR^m$
without explicit knowledge of $\varphi$. Thus
if one uses this edge matrix  to construct a MST over $\calY_n$
its length function will approximate
$L^{\mR^m}_\gamma(\varphi^{-1}(\calY_n))$ and we can invoke Thm.
\ref{T:ManifoldGraph} to characterize its asymptotic convergence
properties. As the edge matrix will contain approximations to the
geodesic distances between pairs of points $(\calY_i,\calY_j)$
this graph will be called a {\it geodesic} MST (GMST).

More specifically, assume that the embedding of $\mM$ is isometric
(conformal) and denote by $\calE_{\mM}$ the edge matrix
$\calE_{\mM}$ over the points $\calY_n$  constructed by the ISOMAP
(C-ISOMAP) algorithm \cite{Bernstein&etal:00,Silva&Tenenbaum:03}
as described in Table \ref{Tab:DistancesEstimate}.  Define the
{\it geodesic} MST $T$ as the minimal graph over $\calY_n$ whose
length is:
\be L^{\mM}_{\gamma}(\calY_n) = \min_{T \in \calT_n} \sum_{e\in T}
|e|_{\mM}^\gamma \label{eq:Liso} , \ee
where $|e|_{\mM}$ ranges over the $n^2$ entries $|e_{ij}|_{\mM}$
of the edge matrix $\calE_{\mM}$ computed by ISOMAP (C-ISOMAP).

The following is the principal theoretical result of this paper and is a simple consequences of %
Thm.  \ref{T:ManifoldGraph}.%
\begin{theorems} \label{T:ConformalGraph1}
Let $\mM$ be a smooth $m$-dimensional manifold embedded in
$[0,1]^d$ through a conformal map $\varphi : \mR^m \mapsto \mM$.
Let $2 \leq m \leq d$ and $0< \gamma < m$. Suppose that $\Vc{Y}_1,
\ldots, \Vc{Y}_n$ are i.i.d.\ random vectors on $\mM$ with common
density $f$ w.r.t.\ Lebesgue measure $\mu_{\mM}$ on $\mM$.  Assume
that each of the edge lengths $|e_{ij}|_{\mM}$ in the edge matrix
$\calE_{\mM}$ converge a.s. to
$|\varphi^{-1}(\Vc{Y}_i)-\varphi^{-1}(\Vc{Y}_j)|_2$ as $n
\rightarrow \infty$.
Then, the length functional of the GMST  satisfies
\be && \lim_{n \to \infty}
L^{\mM}_\gamma(\calY_n)/n^{(d^{'}-\gamma)/d^{'}} 
        \to \label{eq:Linf3}\\
&&\left\{ \begin{array}{cc} \infty , & d^{'} < m \\ \\
        \beta_{m} \int_{\mM}
        f^\alpha(\Vc{y}) \; g^{-\gamma/d}(\varphi^{-1}(\Vc{y}))
        \, \mu_{\mM}(d\Vc{y}) , & d^{'} = m \\ \\
        0, & d^{'} > m \end{array} \right.  
\nonumber
\ee%
(a.s.) where $\alpha=(m-\gamma)/m$ and $g(\Vc{x})\defined \sqrt{\det
\left( \tensor{J}_\varphi^T \tensor{J}_\varphi \right)}$.
Furthermore, the mean
$E[L^{\mM}_\gamma(\calY_n)]/n^{(d^{'}-\gamma)/d^{'}}$ converges to
the same limit.
\end{theorems}

\noindent{\it Proof of Thm. \ref{T:ConformalGraph1}}:

First express the normalized length functional
$L^{\mM}_\gamma(\calY_n)/n^{(d^{'}-\gamma)/d^{'}}$ as
\ben
L^{\mM}_\gamma(\calY_n)/n^{(d^{'}-\gamma)/d^{'}}&=&L^{\mR^m}_\gamma(\varphi^{-1}(\calY_n))/n^{(d^{'}-\gamma)/d^{'}}
\; \\
&&\hspace{0.2in} \cdot
\left[L^{\mM}_{\gamma}(\calY_n)/L^{\mR^m}_{\gamma}(\varphi^{-1}(\calY_n))\right].
\een
By Thm. \ref{T:ManifoldGraph} the first factor on the right
converges (a.s.) to the the limit (\ref{eq:Linf2}). Since the
edges lengths used to construct $L^{\mM}_\gamma(\calY_n)$ converge
a.s. to the edge lengths used to construct $
L^{\mR^m}_\gamma(\varphi^{-1}(\calY_n))$ the term in brackets
converges (a.s.) to 1.  Hence the normalized length functional
$L^{\mM}_\gamma(\calY_n)/n^{(d^{'}-\gamma)/d^{'}}$ converges
(a.s.) to the the limit (\ref{eq:Linf2}). By identifying
$(\alpha-1)=-\gamma/d$, $\Vc{x}=\varphi^{-1}(\Vc{y})$ and $\det
\left( \tensor{J}_\varphi^T \tensor{J}_\varphi \right) =
g(\varphi^{-1}(\Vc{y}))$, for $d^{'}=m$ the integrand on the right
of the limit (\ref{eq:Linf2}) is equivalent to:
$$ f^\alpha(\Vc{y})\;\left[ \det \left( \tensor{J}_\varphi^T \tensor{J}_\varphi
\right) \right]^{\frac{\alpha-1}{2}}= f^\alpha(\Vc{y})\;\left[
g(\varphi^{-1}(\Vc{y})) \right]^{-\frac{\gamma}{2d}} .$$
\qed

If $m>2$, as the parameter $d^{'}$ is increased  from $2$ to
$\infty$ the limit (\ref{eq:Linf3}) in Thm.
\ref{T:ConformalGraph1} transitions from infinity to a finite
limit and finally to zero over three consecutive steps
$d^{'}=m-1,m,m+1$. As $d^{'}$ indexes the rate constant
$n^{(d^{'}-\gamma)/d^{'}}$ of the length functional
$L^{\mM}_{\gamma}(\calY_n)$, this abrupt transition suggests that
the intrinsic dimension $m$ and the intrinsic entropy might be
easily estimated by investigating the convergence rate of the
GMST's length functional. This observation is the basis for the
estimation algorithm introduced in the next section.

We now specialize Theorem \ref{T:ConformalGraph1} to the following
cases of interest.

\subsubsection{Isometric Imbeddings}

In the case that $\varphi$ defines an isometric imbedding the
ISOMAP algorithm is asymptotically able to recover
 the true Euclidean distances between the points in $\calX_n=\varphi^{-1}(\calY_n)$. Thus the
 assumption of Thm. \ref{T:ConformalGraph1} is satisfied.
 Furthermore,
$\tensor{J}_\varphi^T \tensor{J}_\varphi = \tensor{I}_{ m}$. Thus,
for example, when ${L}^{\mM}_\gamma(\calY_n)$ is the length of the
geodesic MST constructed on the edge matrix generated by the
ISOMAP algorithm, the limit (\ref{eq:Linf3}) holds with the
$d^{'}=m$ limit replaced by
$$\beta_{m} \; \int_{\mM}
        f^\alpha(\Vc{y})
\, \mu_{\mM}(d\Vc{y}). $$ Furthermore, $m/\gamma
\log\left(\hat{L}_\gamma^{\mM}(\calY_n)/n^{(m-\gamma)/m}-\log\beta_m\right)$
converges a.s. to the intrinsic entropy (\ref{eq:intent}).

\subsubsection{Isometric Imbeddings with Contraction/Expansion}

In the case that $\varphi$ defines an isometric imbedding with
contraction or expansion the C-ISOMAP algorithm is able to recover
the true Euclidean distances between points in $\calX_n$.
Furthermore, $\tensor{J}_\varphi^T \tensor{J}_\varphi = c \,
\tensor{I}_{m}$ where $c$ is a constant. Thus, when
${L}^{\mM}_\gamma(\calY_n)$ is the length of the geodesic MST
constructed on the edge matrix generated by the C-ISOMAP algorithm
the limit (\ref{eq:Linf3}) holds with the $d^{'}=m$ limit replaced
by
$$\beta_{m} c^{-\gamma/2}\; \int_{\mM}
        f^\alpha(\Vc{y})
\, \mu_{\mM}(d\Vc{y}). $$
Now $m/\gamma \log
\left(\hat{L}_\gamma^{\mM}(\calY_n)/n^{(m-\gamma)/m}-\log
\beta_m\right)$ converges a.s. up to an unknown additive  constant
$-\gamma/2\log c$ to the intrinsic entropy (\ref{eq:intent}). We
point out that in many signal processing applications (e.g.\ image
registration) a constant bias on the entropy estimate does not
pose a problem since an estimate of the relative magnitude of the
entropy functional is all that is required.

\subsubsection{Non-isometric Imbeddings Defined by Conformal Mappings}

In the case that $\varphi$ is  a general (non-isometric) conformal
mapping the C-ISOMAP algorithm is once again able to recover the
true Euclidean distances between points in $\calX_n$. Furthermore,
$\tensor{J}_\varphi^T \tensor{J}_\varphi = c(\Vc{x}) \,
\tensor{I}_{m}$. Thus, when ${L}^{\mM}_\gamma(\calY_n)$ is the
length of the geodesic MST constructed on the edge matrix
generated by the C-ISOMAP algorithm, the limit (\ref{eq:Linf3})
holds with the $d^{'}=m$ limit replaced by
$$\beta_{m} \int_{\mM}
        f^\alpha(\Vc{y})  \; c^{-\gamma/2}(\varphi^{-1}(\Vc{y}))
\, \mu_{\mM}(d\Vc{y}). $$
In this case $m/\gamma \log
\left(\hat{L}_\gamma^{\mM}(\calY_n)/n^{(m-\gamma)/m}-\log\beta_m\right)$
converges a.s. up to an additive  constant to the weighted
intrinsic entropy
 \ben
 \label{eq:went}
 \frac{1}{1-\alpha} \log \int_{\mM} f^\alpha(\Vc{y}) \; c^{-\gamma/2}(\varphi^{-1}(\Vc{y})) \,
 \mu_{\mM}(\Vc{y}) \ . \een
 The weighted $\alpha$-entropy %
 is a ``version" of the standard unweighted
$\alpha$-entropy
    $H^{\mM}_\alpha(f)$ which is
    ``tilted" by the space-varying volume element of $\mM$.  This
    unknown weighting makes it impossible to estimate the intrinsic unweighted $\alpha$-entropy.
    However, as can
    be seen from the discussion in the next section, as the rate exponent of the
    GMST length depends on $m$ we
    can still perform dimension estimation in this case.

\subsubsection{Non-conformal Diffeomorphic Imbeddings}

When $\varphi$ defines a general diffeomorphic embedding a result
analogous to Thm. \ref{T:ManifoldGraph} easily follows giving an
identical limiting relation to (\ref{eq:Linf2}) except that
${L}^{\mM}_\gamma(\calY_n)/n^{(d^{'}-\gamma)/d^{'}} $ converges to
$$\beta_{m} \int_{\mM}
        f^\alpha(\Vc{y})\;
        \left[ \det \left( \tensor{J}_\varphi^T \tensor{J}_\varphi \right) \right]^{-\gamma/2d}
\, \mu_{\mM}(d\Vc{y}), $$
when $d^{'}=m$.
However,   without an extension of the C-ISOMAP algorithm that can
provably learn the Euclidean distances between the points
$\calX_n$ in the parametrization space,  Thm.
\ref{T:ConformalGraph1} is not applicable. To the best of our
knowledge such an extension of C-ISOMAP does not yet exist.

\section{GMST Algorithm}
\label{S:GMST}

{\tiny \begin{table}
\noindent{\tt Initialize: Using entire database of \\signals 
$\calY_n$ construct geodesic distance \\ matrix  $\calE_{\mM}$ using
ISOMAP or C-ISOMAP.}

\noindent{\tt Select parameters: \\ $p_0$, $p_1$ $(p_0< p_1\leq n)$,
and $N$ $(N>0)$}

\bt XX\= XX \= XX \= \kill
{\tt for $p=p_1, \ldots , p_Q$}
\\
\>{$\ol{L}=0$}
\\
\> {\tt for $N'=1,\ldots, N$}
\\
\> \> {\tt Randomly select a subset of $p$ signals $\calY_p$ from
$\calY_n$}
\\
\> \> {\tt Compute geodesic MST length $L_p$ over $\calY_p$} 
\\ \> \>
$\ol{L}=\ol{L}+L_p$
\\
\> {\tt end for}
\\
\> {\tt Compute sample average geodesic MST length }
\\ \>
$\hat{E}[{L}_\gamma^{\mM}(\calY_p)]=\ol{L}/N$
\\
{\tt end for}
\\
{\tt Estimate $m$ and $H_\alpha^{\mM}(f)$ from
$\{\hat{E}[L_\gamma^{\mM}(\calY_p)]\}_{p=p_1}^{p_Q}$ }
\et
 \caption{GMST resampling algorithm for estimating
intrinsic dimension $m$ and intrinsic entropy $H_\alpha^{\mM}$.}
\label{A:algp}
\end{table}
}

Now that we have characterized the asymptotic limit
(\ref{eq:Linf3}) of the length function of the GMST we here apply
this theory to jointly estimate entropy and dimension. The key is
to notice that the rate of convergence is strongly dependent on
$m$ while the rate constant in the convergent limit is equal to
the intrinsic $\alpha$-entropy. We use this strong rate dependence
as a motivation for a simple estimator of $m$. Throughout we
assume that the geodesic minimal graph length
$L^{\mM}_\gamma(\calY_n)$ is determined from an edge matrix
$\calE_{\mM}$ that satisfies the assumption of Thm.
\ref{T:ConformalGraph1}, e.g.,  obtained using ISOMAP or C-ISOMAP.
We set the edge power weighting in $L^{\mM}_\gamma(\calY_n)$ to
$\gamma=1$ and assume that $m \geq 2$. This guarantees that
$L^{\mM}_\gamma(\calY_n)/n^{(d^{'}-\gamma)/d^{'}}$ has a non-zero
finite convergent limit for $d^{'}=m$.  Next define $l_n=\log
L^{\mM}_\gamma(\calY_n)$. According to (\ref{eq:Linf3}) $l_n$ has
the following approximation
\be l_n= a \; \log n + b + \epsilon_n\label{E:addmod},\ee
where
\be a&=&(m-\gamma)/m,
\nonumber \\
b&=&\log \beta_m + \gamma/m \; H_{\alpha}^{\mM}(f),
\label{E:abdef} \ee
$\alpha=(m-\gamma)/m$ and $\epsilon_n$ is an error residual that
goes to zero a.s. as $n \rightarrow \infty$.

The additive model (\ref{E:addmod}) could be the basis for many
different methods for estimation of  $m$ and $H$. For example, we
could invoke a central limit theorem on the MST length functional
\cite{Alexander:AnnAppProb96} to motivate a Gaussian approximate
to $\epsilon_n$ and apply maximum likelihood principles. However,
in this paper we adopt a simpler non-parametric least squares
strategy which is based on resampling from the population
$\calY_n$ of available points in $\mM$. The algorithm is
summarized in Table \ref{A:algp}. Specifically, let $p_1, \ldots,
p_Q$, $1 \leq p_1 < \ldots, <p_Q \leq n$, be $Q$ integers and let
$N$ be an integer that satisfies $N/n=\rho$ for some fixed $\rho
\in (0,1]$. For each value of $p\in \{p_1, \ldots, p_Q\}$ generate
$N$ independent samples $\calY_p^{j}$, $j =1, \ldots, N$ and from
these samples compute the empirical mean of the GMST length
functionals $\ol{L}_p = N^{-1} \sum_{j=1}^N
L^{\mM}_\gamma(\calY_p^j)$.  Defining $\ol{\Vc{l}}=[\log
\ol{L}_{p_1} , \ldots, \log \ol{L}_{p_1}]^T$, and motivated by
(\ref{E:addmod})  we write down the linear vector model
\be \ol{\Vc{l}}=A \left[\begin{array}{c} a \\ b \end{array}\right]
+ {\boldmath{\epsilon}} \label{E:lls}\ee
where
$$
A=\left[\begin{array}{ccc} \log p_1 & \ldots & \log p_Q \\ 1 &
\ldots & 1 \end{array} \right]^T.
$$
Expressing $a$ and $b$ explicitly as functions of $m$ and
$H_\alpha$ via (\ref{E:abdef}), the dimension and entropy
quantities could be estimated using a combination of non-linear
least squares (NLLS) and integer programming.
Instead we take a simpler method-of-moments (MOM) approach in
which we use (\ref{E:lls}) to solve for the linear least squares
(LLS) estimates $\hat{a},\hat{b}$ of $a,b$ followed by inversion
of the relations (\ref{E:abdef}). After making a simple large $n$
approximation, this approach yields the following estimates:
\ben \hat{m}&=& \lfloor\gamma/(1-\hat{a})\rfloor \\
\hat{H}_\alpha^{\mM}&=& \frac{\hat{m}}{\gamma}\left( \hat{b}- \log
\beta_{\hat{m}}\right) .\een
It is easily shown that the law of large numbers and Thm.
\ref{T:ConformalGraph1} imply that this estimator is consistent as
$n \rightarrow \infty$.  We omit the details.

A word about determination of the sequence of constants
$\{\beta_m\}_{m}$ is in order. First of all, in the large $n$
regime for which the above estimates were derived, $\beta_m$ is
not required for the dimension estimator. $\beta_m$ is the limit
of the normalized length functional of the Euclidean MST for a
uniform distribution on the unit cube $[0,1]^m$.
Closed form expressions are not available but several
approximations and bounds can be used in various regimes of $m$
\cite{Yukich:98,Avram&Bertsimas:AnnAppProb92}. Another possibility
is to determine $\beta_m$ by simulation of the Euclidean MST
length on the $m$-dimensional cube for uniform random samples. In
our simulations, described below, we have used the large $m$
approximation of Bertsimas and van Ryzin
\cite{Bertsimas&vanRyzin:ORL90}: $\log \beta_m \approx \gamma/2 \;
\log(m/2\pi e)$.

Before turning to the application we briefly discuss computational
issues.  We have developed a custom implementation of the MST
algorithm which is a modification of Kruskal's algorithm
\cite{Heemuchwala&etal:SP03}. This implementation implements an
efficient disk radius algorithm to restrict the search space
yielding substantial runtime speedup. This has allowed us to
routinely implement the MST on tens of thousands of points.

\section{Application}
\label{S:Application}

We performed several preliminary validation tests of the GMST
estimator on simulated data including: a linear manifold and the
swiss roll manifold investigated in
\cite{Tenenbaum&etal:Science00}. Due to space limitations we will
not present results from these validation tests.  Rather we will
present a very simple example to illustrate the applicability of
GMST intrinsic dimension and entropy estimates. For this purpose
we investigated a set of  black-and-white images of several
individuals taken from the  Yale Face Database B \cite{GeBeKr01}.
This is a publicly available database containing face images of 10
subjects with 585 different viewing conditions for each subject.
These consist of 9 poses and 65 illumination conditions (including
ambient lighting).  The images were taken against a fixed
background which we did not bother to segment out. We think this
is justified  since any fixed structures throughout the  images
would not change the intrinsic dimension or the intrinsic entropy
of the dataset. We randomly selected 3 individuals from this data
base and subsampled each person's face images down to a $64 \times
64$ pixel image. The pixels in each of the images were
lexicographically reordered into vectors residing in a 4096
dimensional space.

We studied the dimension and entropy of each person's face  as
follows. We first generated the Euclidean nearest neighbor graph
$\Gg$ used by ISOMAP in Step 1 (see Table
\ref{Tab:DistancesEstimate}) for each of the three sets of  585
images. We then investigated the trajectory of the mean GMST as a
function of $n$ for each person's face folio.  Specifically $26
\times 25$ random samples (with replacement) were selected to form
26 resampled face subsets of sizes ranging from $100$ to $585$,
respectively. Step 2 of the ISOMAP algorithm was then implemented
on each sample to generate 650 different edge matrices.
Subsequently the GMST was computed from each of these edge
matrices and for each of the 26 folio sizes the 25 resampled GMST
length functions were averaged to obtain 3 average GMST length
sequences over $n$. In the GMST implementation the edge exponent
$\gamma$ was fixed at a value of 1.

\begin{figure}[htbp]%
\centering%
\epsfxsize=3in \epsffile{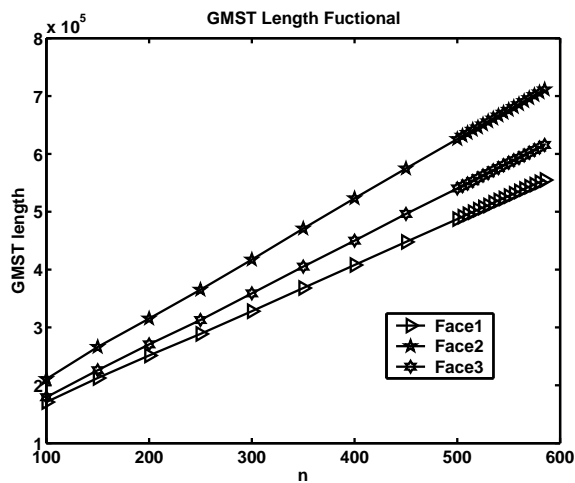} \caption{The average geodesic
MST growth rates for three different face images in the Yale face
database B.}
 \label{F:facecurve}
 \end{figure}

\begin{figure}[htbp]%
\centering%
\epsfxsize=3in \epsffile{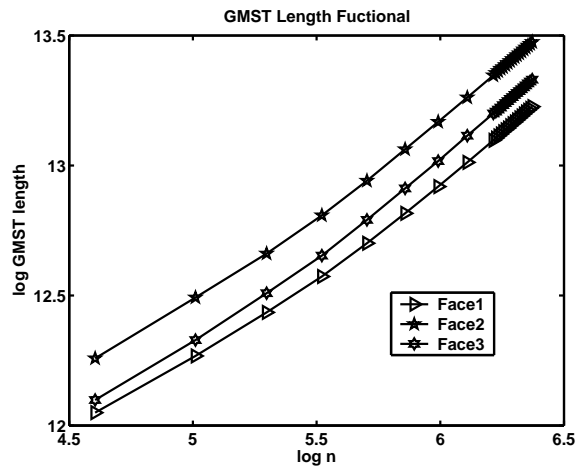} \caption{Log-log plot of Fig.
\ref{F:facecurve}.}\label{F:facecurve2}\end{figure}

\begin{figure}[htbp]%
\centering%
\epsfxsize=3in \epsffile{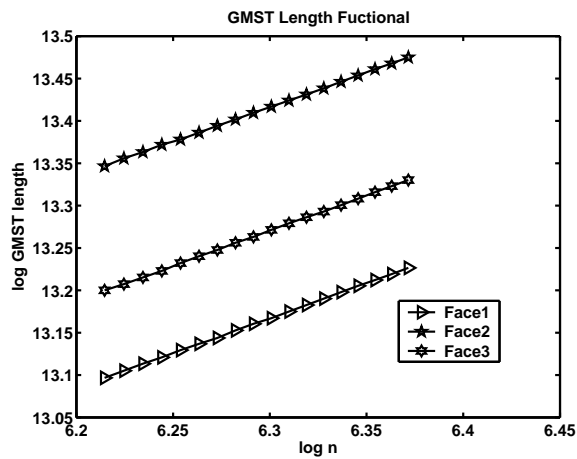}  \caption{Blowup of
Fig. \ref{F:facecurve2} showing linearity of geodesic MST growth
rates for large $n$.}\label{F:facecurve3}\end{figure}

In Fig. \ref{F:facecurve} the sequence of average GMST length
functionals is plotted for each of the three faces. The symbols
denote the locations of the 26 values of $n$ chosen for study and
the corresponding values of the average GMST length. Note that the
average GMST length sequences appear to increase almost linearly
over $n$ for each of the three persons, albeit with different rate
constants. However, after a log-log transformation, shown in Fig.
\ref{F:facecurve2}, it becomes evident that the linear model for
the of the mean GMST length functional  is not valid for small
$n$. Fig. \ref{F:facecurve3} is a blowup of Fig.
\ref{F:facecurve2} for $n\geq 500$ and experimentally confirms the
large-$n$ linear behavior predicted by Thm.
\ref{T:ConformalGraph1} and supports the validity of the linear
model (\ref{E:addmod}).

Using the average GMST length sequences we next estimated slope
and intercept parameters $a,b$ of the linear model and implemented
the MOM estimator of dimension and entropy as described in the
previous section. Only the range $n>500$ was used in fitting the
linear model. The MOM estimator of $m$ was rounded to the nearest
integer and the parameter $\beta_m$ was estimated by the large $m$
approximation \cite{Bertsimas&vanRyzin:ORL90}. The results are
summarized in Table \ref{T:faces}. As a result of this procedure
the estimated face dimension $m$ was observed to vary between 5
and 6 for each of the individuals. The intrinsic entropy estimate
expressed in log base 2 was concentrated around 70 bits. Note that
as $\alpha=(m-1)/m$  is close to one for these estimated values of
$m$ the estimates of $\alpha$-entropy are expected to be close to
the Shannon entropy. These entropy estimates suggest that one
should be able to get away with a model incorporating at most 6
parameters to describe the range 585 poses and illuminations of
any of the three faces. An MDS ISOMAP analysis of the same three
faces gave slightly higher estimates of dimension, varying between
6 and 7.

\begin{table}
\centering
\begin{tabular}{|c||c|c|c|}
  \hline
  & Face1 & Face2 & Face3 \\
  \hline\hline
  $\hat{m}$ & 6 & 5 & 6 \\
  \hline
  $\hat{H}$ (bits) & 70.4 & 68.8 & 73.8 \\
  \hline
\end{tabular}
\caption{Dimension estimates $\hat{m}$ and entropy estimates
$\hat{H}$  for three faces in the Yale Face Database B.}
\label{T:faces}
\end{table}

\section{Conclusion}

We have presented a novel method for intrinsic dimension
estimation and entropy estimation on smooth domain manifolds. With
regards to intrinsic dimension estimation, the method proposed has
two main advantages. First, it is global in the sense that tyhe
MST is constructed over the entire and we thus  avoid local
linearizations. Second, unlike previous methods it simple to
implement and does not require tuning any user-defined parameters
such as eigenvalue thresholds or sizes of local neighborhoods. The
GMST methods described in this paper are currently being applied
to a large number of dimension reduction and entropy
characterization problems including: gene clustering in
bioinformatics, Internet traffic analysis, lung nodule
classification, and radar signature analysis.

\bc {\bf Acknowledgments} \ec

The authors acknowledge and thank the creators of Yale Face
Database B for making their face image data publicly available
({\small
\verb|cvc.yale.edu/projects/yalefacesB/yalefacesB.html|}). The
authors would also like to thank Huzefa Neemuchwala and Arpit
Almal for their help in acquiring and processing these face
images. This work was partially supported by the DARPA MURI
program under ARO contract DAAD19-02-1-0262 and by the NIH Cancer
Institute under grant 1PO1 CA87634-01.  The authors can be contacted by email 
via {\verb|jcosta,hero@eecs.umich.edu|}.

\end{document}